DesignCon® 2026
WHERE THE CHIP MEETS THE BOARD

# AI-driven Thermal-aware Data Center Capacity Planning

Yixing Li, Cadence Design Systems
yixingli@cadence.com

Mark Fenton, Cadence Design Systems
mfenton@cadence.com

Matthew Kaufeler, Cadence Design Systems
kaufeler@cadence.com

Ka Ming Leung, Cadence Design Systems
kleung@cadence.com

Xin Ai, Cadence Design Systems
nathanai@cadence.com

Zhiyu Zeng, Cadence Design Systems
zzeng@cadence.com

## Abstract

The emerging of large language models (LLMs) has posed significant challenges to the thermal management of data center. Intense GPU computation for LLMs results in localized hotspots. Moreover, spiking thermal loads during training and inference bursts make real-time cooling response more difficult to predict and control. Thermal-aware capacity planning of data center requires massive expensive high-fidelity CFD simulations. AI models can perform real-time prediction for unseen designs. However, existing works either have large prediction error, or have over-simplified assumptions for data center operations.

This work presents an AI-driven framework that can perform thermal-aware capacity planning for a real-world data center in seconds. The embedded AI model learns from numerous key parameters (rack power, server power, server placement, HVAC settings etc.), and provides temperature prediction within milliseconds. This AI model is tested against high-fidelity CFD simulations, and results show that for unseen data center designs, model can achieve high accuracy with 10000X speedup. Driven by the AI model, the authors design the thermal-aware capacity planning framework. This framework can help data center designers and operators instantaneously optimize both workload distribution and HVAC cooling efficiency.

## Author(s) Biography

Dr. Yixing Li earned his Ph.D. from the Georgia Institute of Technology in 2019 and is a Senior Principal Software Engineer at Cadence. With deep expertise in CFD and FEA, he combines strong foundations in physics and engineering with modern AI. He has led and productized multiple key AI initiatives, delivering production grade systems across Cadence technologies.

Mark Fenton has over 20 years of experience applying computational fluid dynamics to data center design and operations. He has helped develop and deploy advanced technical software in a rapidly growing industry. As Product Engineering Director for the Cadence Reality Digital Twin Platform, he drives strategic partnerships and delivers solutions for enterprise, hyperscale, colocation, and AI Factory customers.

Matthew is a Principal Product Engineer for Cadence's Reality DC Digital twin software for data centers. Following a degree in Mechanical Engineering at Imperial College London, he worked in R&D, product development and technical sales at cooling hardware company before moving to Cadence in an application engineering role.

Ka Ming Leung, Software Engineering Group Director at Cadence Design Systems, leads engineering initiatives in London and oversees the RealityDC and CelsiusEC product lines. With 18+ years' experience, he architected and initiated the development of RealityDC Insight, bridging business goals with technical execution while driving innovation.

Dr. Xin Ai, Ph.D from University of Washington, serves as Software Engineering Group Director at Cadence Design Systems, where he has led the Celsius RD team. He initiated and drove the development, launch, and ongoing enhancement of the Celsius Thermal Solver and Celsius Studio, shaping them into core components of Cadence's thermal and multi physics solutions.

Dr. Zhiyu Zeng received his Ph.D. from Texas A&M University in 2011 and joined Cadence in 2012, where he is now Sr. Software Engineering Group Director. He has led major RD teams across Voltus, Celsius, and Reality, driving innovations such as Voltus XP, Voltus InsightAI, Celsius Studio, and the Reality Digital Twin Platform. He currently focuses on GPU acceleration and AI at HSA. His contributions have been recognized with the EE Awards Asia Innovation Award and the Cadence Achievement Award.

## 1. Introduction

The rapid rise of large-scale large language model (LLM) workloads has dramatically increased power demands of modern data centers [1]. Dense GPU clusters generate concentrated heat fluxes, creating localized hotspots and transient thermal spikes that challenge existing cooling infrastructures. Managing such fluctuating and spatially non-uniform thermal loads requires fast and accurate prediction of temperature distribution and real-time adjustment of workload and cooling strategies.

Traditional computational fluid dynamics (CFD) analysis provides high-fidelity simulation results but is extremely resource-intensive, for both computing power and time [2], [3]. Full-scale high-fidelity simulation of a single data center configuration may require hours or even days to complete, making it impractical for real-time thermal management or rapid design-space exploration.

To enable temperature-field prediction with significantly reduced turnaround time and computational cost, researchers have explored a wide range of techniques that approximate the behavior of full CFD solvers. Black box model use only a few states and act as input output mappings derived from physical laws or empirical data. Gray box models introduce partial physical characterization and use system identification to estimate unknown parameters, offering better fidelity while remaining lightweight. Surrogate models increase the number of states moderately and embed additional geometric or physical structure so they can reproduce meaningful internal temperature or flow distributions without the cost of full CFD or CHT analysis [2], [4], [5], [6]. Recently, with the rapid evolution of machine learning methods, AI based approaches have been integrated with surrogate models, where neural networks augment simplified physics or approximate flow and temperature fields directly. These hybrid methods capture nonlinear thermal interactions that are difficult to model analytically and provide full field predictions at far lower cost than CFD, making them well suited for data center thermal mapping and real time analysis [7], [8]. On the other hand, while these methods are conceptually appealing, they rely on simplified representations of airflow and heat transfer and are typically tailored to a specific data center configuration, which limits their generalizability.

A notable advancement in AI-driven data center surrogate model was introduced by a work from HPE [9], which applied a 3D CNN-based CFD surrogate model and further integrated the data-driven model to a GA optimization framework. Their framework successfully reduced both maximum normalized temperature and cooling energy consumption. The predictive accuracy of the AI surrogate remained limited, with reported maximum absolute errors ranging from 4 °C to 20 °C across different 3D U-Net variants.

In the present study, the authors noticed that data center operators often prioritize a subset of locations, particularly cabinet inlet, which directly indicate cooling sufficiency and hotspot risk. Motivated by these insights, the present work specially emphasizes on accurate prediction of cabinet inlet thermal conditions, which represent the most operationally critical points for data center designers and operators.

The innovations of this work are two-fold. From a methodological standpoint, we develop a data-driven AI model capable of predicting cabinet inlet temperatures with high accuracy (MAE < 0.25 °C) and achieving more than 10,000× speedup over full CFD simulation. We demonstrate its applicability in real-world capacity planning scenarios. In addition, we integrate this modeling framework into the Reality Digital Twin platform, creating a fully automated pipeline for data generation, model training, and real-time inference. This integration bridges the gap between academic surrogate-model research and industrial deployment, enabling engineers to apply the framework directly to practical data center challenges without rebuilding the workflow. Users may leverage the provided baseline model or train their own specialized AI surrogate tailored to specific design requirements. By combining flexible front-end visualization with ultra-fast back-end inference, the system supports drag-and-drop interaction with instantaneous temperature prediction. Building on this capability, we further implement a rapid decision engine that evaluates workload or equipment placement using an AI-driven strategy rather than traditional heuristic rules.

## 2. Methodology and System Setup

### 2.1 Cadence Reality DC Design and Reality Digital Twin Platform

Cadence Reality DC Design (formerly Future Facilities 6SigmaDCX) enables engineers to design the next generation of data centers, with physics-based simulation powered by computational fluid dynamics (CFD). It allows modeling of full data-center geometries for all variations of layouts, from chip to chiller with multi-core CPU and GPU solving capacity. The software enables engineers to construct accurate 3D digital-twin models, explore multiple design configurations, and assess failure scenarios to optimize resilience, capacity, and energy efficiency. It supports a wide range of cooling strategies, including raised floors, fan walls, evaporative cooling, and liquid cooling. and provides industry-standard metrics such as ASHRAE compliance, PUE, and Green Grid PI. Reality DC Design integrates a 1D-3D co-simulation framework for liquid-cooling and airflow networks, models modern control systems, and visualizes power networks from UPS to cabinet level. The tool also incorporates environmental modeling (wind, humidity, solar gain) and transient simulation capabilities to study the impact of dynamic IT loads or infrastructure failures over time. With access to an extensive library of intelligent vendor items and automated reporting tools, Cadence Reality DC Design streamlines modeling, solving, and documentation, enabling users to efficiently develop robust and sustainable data-center designs.

Cadence Reality DC Digital Twin [10] uses physics-based simulation to provide the visibility and foresight necessary to take control of data center performance. As shown in Figure 1, this platform combines detailed digital twin models with a powerful CFD solver. The digital twin model is calibrated to reflect its physical counterpart accurately. By combining a detailed digital-twin model with a high-fidelity CFD solver, the platform provides predictive visibility into airflow, cooling behavior, power distribution, and operational risks. The digital twin can be continuously synchronized with live building-management (BMS) and data-center-infrastructure (DCIM) systems, ensuring that it accurately reflects real-world conditions. It supports capacity visualization across power, space, cooling, network, and weight; evaluates deployment options and failure scenarios; and models a wide range of cooling strategies, including raised floors, fan walls, adiabatic systems, and liquid cooling. The platform enables automated simulation scheduling, reporting, and workflow integration, and provides operational insights through dashboards, power-network diagrams, SLA and ASHRAE compliance views, and sustainability metrics such as PUE and carbon usage. Together, these capabilities allow data-center teams to optimize performance, reduce deployment risk, and make informed, energy-efficient decisions throughout the data-center lifecycle.

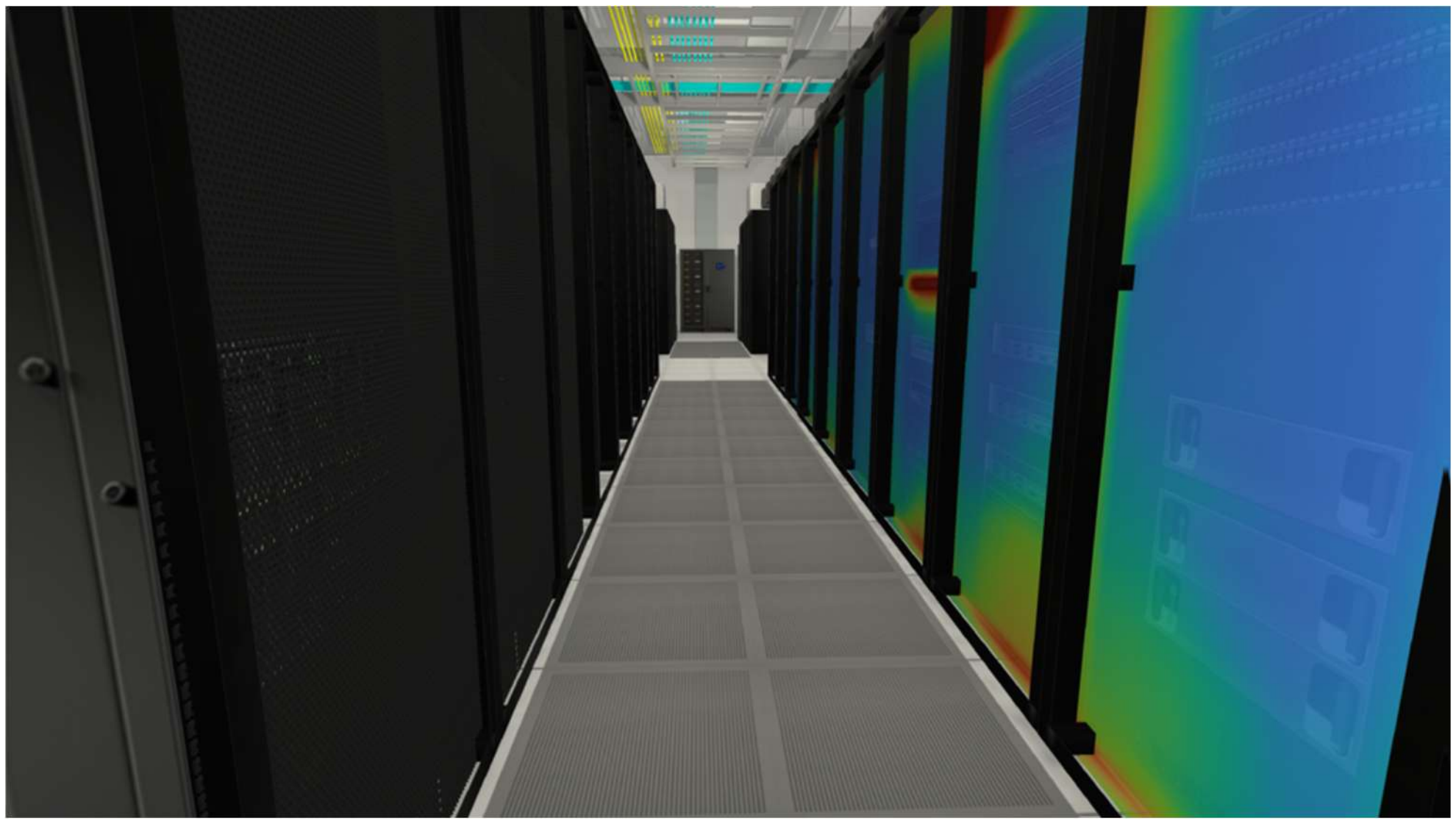

Figure 1 Cadence Reality DC Digital Twin showing simulation results of the cabinet thermal temperature plane

Cadence Reality DC Digital Twin extends the capabilities of DC Design by transforming high-fidelity CFD modeling into a continuously updated, operationally aware decision-support platform. It provides advanced capabilities such as capacity and resource-utilization visualization, automated simulation scheduling, deployment guidance, SLA/ASHRAE and sustainability metrics, and integration with operational workflows. As a result, the Digital Twin enables ongoing performance optimization and risk reduction across the entire data-center lifecycle, offering a more comprehensive, automated, and operationally meaningful solution than design-only simulation environments.

### 2.2 Data acquisition

To train the proposed AI-driven surrogate model, it is necessary to generate a comprehensive dataset that captures the thermal behavior of a wide range of data center configurations and operating conditions. Raw training data were generated by conducting a series of data-center simulations using the Cadence Reality platform. The data-collection process was automated through Commander, a built-in command-line utility that enables users to execute scripted sequences of operations within Cadence Reality.

Prior to running simulations, it is essential to identify the input parameters to be varied and to define the corresponding parameter ranges. These parameters can span multiple hierarchical levels of the data center model and are customizable depending on the specific design. At the rack level, parameters may include the number of cabinets, cabinet dimensions, and the presence of empty or reserved spaces within the layout. At the cooling-unit level, relevant parameters include the ACU setpoint and volumetric flow rate. At the IT-equipment level, parameters include equipment placement, equipment geometry, and power allocation. Once the parameter set and ranges are defined, Commander can be configured to automatically generate multiple simulation runs. As a practical guideline, initial datasets typically consist of approximately 500 simulation cases with around ten randomized parameters. As training progresses, additional cases can be generated and incorporated into the dataset to further improve model accuracy and robustness.

Upon completion of data acquisition, all input parameters for each simulation are aggregated as the training inputs. For training outputs, the authors extract the CFD-simulated results (temperature, velocity, pressure) at the cabinet inlet. Cabinet inlet conditions provide the most operationally critical indicators of whether cool air is under-supplied or over-supplied to each rack and IT. Furthermore, when combined with IT-level power loading, inlet temperature is a primary determinant of potential local hotspot formation. Consequently, focusing on cabinet inlet predictions provides the most physically meaningful surrogate-model output for data-center designers and operators.

### 2.4 AI model architecture

To capture the spatially varying thermal behavior represented in CFD simulations, the surrogate model must learn nonlinear mappings between multi-dimensional input fields and the resulting temperature distribution. U-Net is selected for its strong predictive accuracy and its ability to retain fine-scale spatial details. U-Net is a convolutional neural network originally introduced for biomedical image segmentation [11] and is widely used for pixel-level prediction problems. Its encoder–decoder structure with skip connections enables the model to extract global features while preserving fine-scale spatial details.

In our implementation, simulation inputs are organized as a multi-channel tensor, with each channel representing a specific parameter such as cabinet parameters, equipment placement, or HVAC settings. The U-Net processes this multi-channel input and outputs a single channel corresponding to the cabinet inlet temperature field, allowing the model to learn how input parameters jointly influence spatial thermal behavior. Although the simulation environment also provides temperature, velocity, and pressure fields at both cabinet inlet and outlet locations, we focus on predicting the inlet temperature because it is the most operationally critical indicator of cooling sufficiency and hotspot risk. The U-Net processes the multi-channel input and outputs a single channel corresponding to the cabinet inlet temperature distribution, capturing how input conditions jointly influence spatial thermal behavior.

## 3. Case Setup and Results

### 3.1 Case Setup

We will present 2 configurations. The first configuration corresponds to a raised floor data center design, as shown in Figure 2. The room dimensions are 10.8 m by 9 m with a height of 4.5 m, and the layout consists of 4 rows containing a total of 32 cabinets. Each cabinet includes 21 IT equipment units, with each unit having a height of 2U. Four air conditioning units (ACUs) are positioned at the row ends, with two units located on each side. In this arrangement, conditioned air is supplied into the subfloor plenum by the ACUs and delivered to the cold aisle through perforated floor grilles. The cold aisle is fully contained to ensure directed airflow to cabinet inlets. After passing through the IT equipment, the heated exhaust air rises and mixes within the room before returning to the ACUs for recirculation.

The randomization strategy is designed to span a wide range of realistic operating conditions and capture the thermal behavior of the raised floor configuration under diverse load and cooling scenarios. Power values for individual IT units are sampled uniformly between 0 and 10 kW, representing different IT workloads and job profiles. A subset of cabinets and IT units is assigned zero power to emulate idle or partially populated racks commonly observed in production environments. The fan speed of each ACU is varied across simulation cases to represent different airflow delivery strategies and cooling settings, allowing the dataset to reflect a broad spectrum of cooling behaviors.

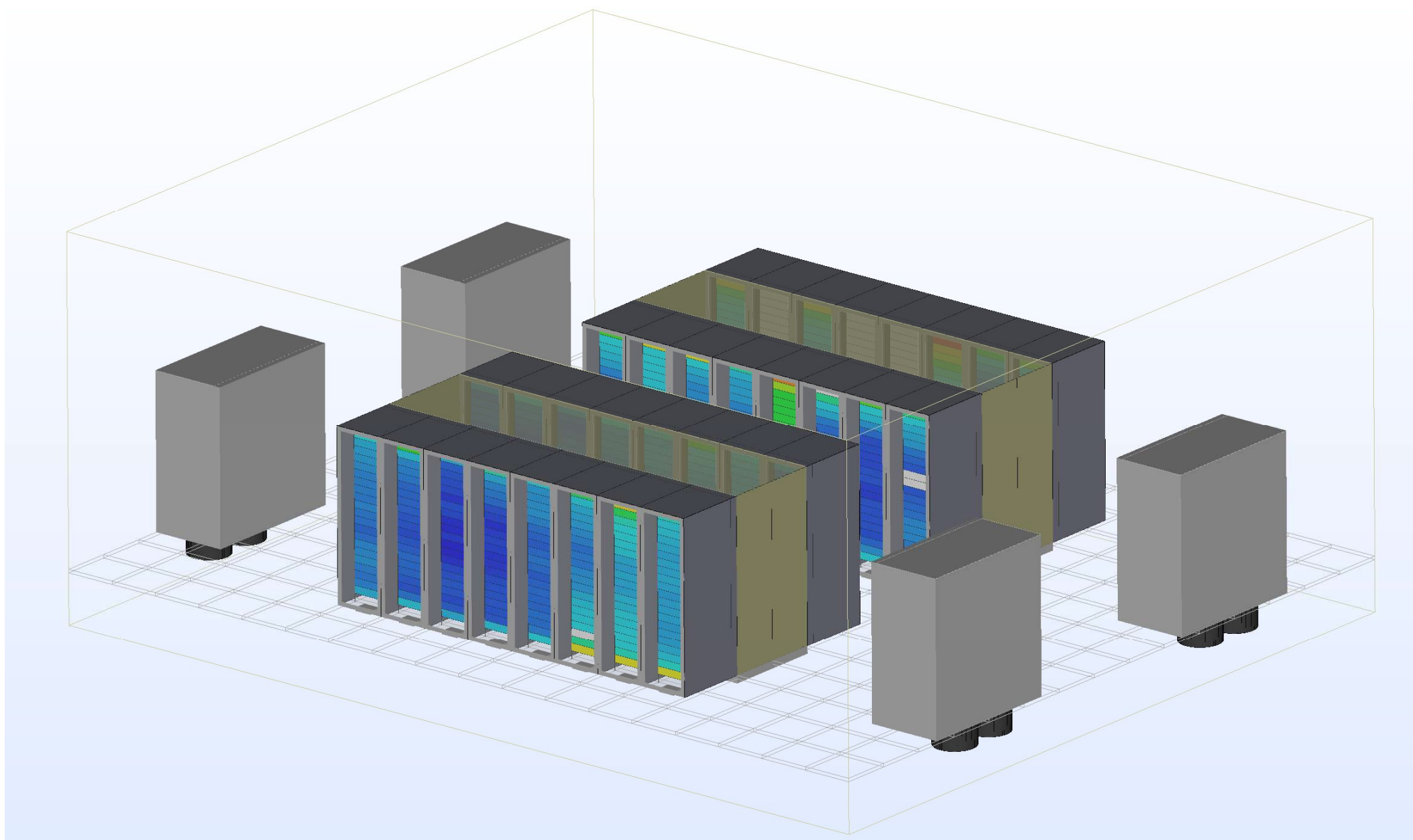

Figure 2 Diagram of raised floor design

### 3.2 AI Prediction Results

After we have collected results of 1k data center simulation cases, the resulting dataset was used to train the U-Net model. Epoch number is 1000. Dataset was split with 85% of training, 10% testing, 5% evaluation, and it takes approximately 5 hours to train the model on four Nvidia A100 GPUs.

Evaluation metrics for the evaluation dataset are summarized in Table 1. We used mean average error and max average error as evaluation metrics. Both mean absolute error and maximum absolute error were computed. Prediction accuracy was assessed at two levels: the cabinet inlet field level, which measures point to point agreement of the predicted temperature distribution, and the cabinet inlet maximum temperature level, which evaluates accuracy of the cabinet level peak temperature as a scalar feature. The model achieved low mean absolute errors for both field prediction and maximum temperature prediction. However, the maximum absolute error for the absolute temperature field remained relatively large, a trend also reported in reference [9].

Inference performance was also quantified. The surrogate model achieved an average inference time of 9 milliseconds per case, yielding to a runtime speedup of approximately 150,000 times compared with full scale CFD simulation, which requires roughly 25 minutes for the same configuration.

| **Metric type** | **Mean AE (C)** | **Max AE (C)** |
|---|---|---|
| Cabinet inlet field prediction | 0.23 | 12.11 |
| Cabinet inlet maximum temperature | 0.41 | 6.79 |

Table 1 Evaluation metrics for cabinet inlet field prediction and cabinet inlet maximum temperature prediction.

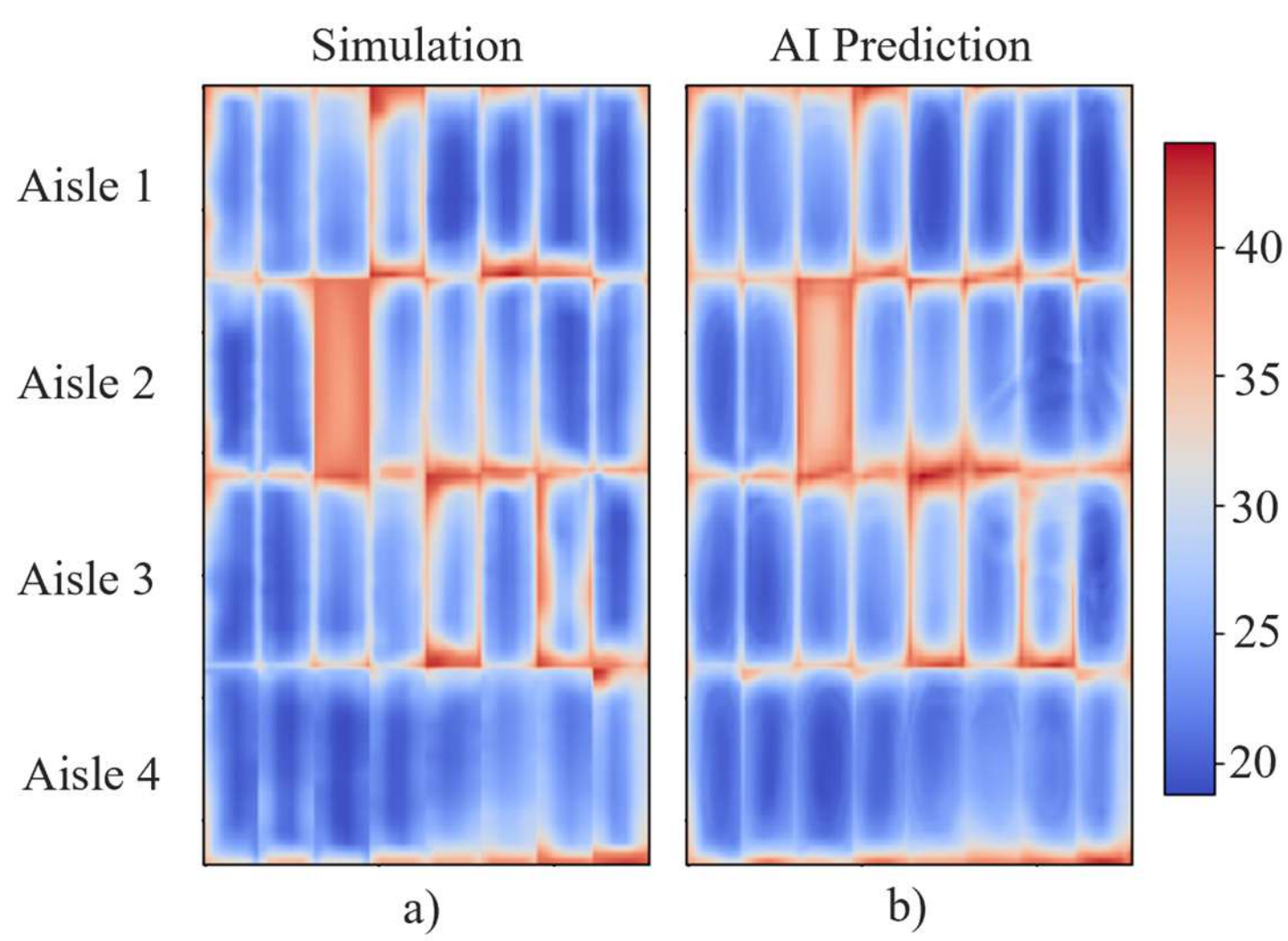

Figure 3 a) Simulation results of the raised floor design at the cabinet inlet, b) AI prediction of the raised floor design at the cabinet inlet

Figure 3 visualizes a side-by-side comparison of simulation results (left figure) and AI prediction (right figure) of the raised floor design at the cabinet inlet. Visualization indicates model's ability to capture high-fidelity temperature distribution at cabinet inlet.

For many practical scenarios, the maximum inlet temperature of each cabinet is often a more critical indicator than the full temperature field. Therefore, the cabinet inlet temperature field was further summarized into a scalar metric representing the maximum inlet temperature for each cabinet, as depicted in Figure 4. Although a small number of cabinets exhibit relatively large prediction errors of approximately 3C, the model consistently preserves the relative ordering of temperatures across cabinets, correctly identifying which cabinets operate at higher temperatures than their neighbors. This property is highly valuable for data center designers and operators for several reasons. Relative temperature ranking is a primary indicator of which cabinets are most thermally stressed, enabling rapid identification of potential hotspots even when the absolute error is nontrivial. Based on the information of which racks are comparatively warmer, operators are able to redistribute and optimize workloads. Therefore, even with modest absolute errors in a few locations, preserving the correct temperature ordering provides practical guidance for operational decisions and thermal risk mitigation.

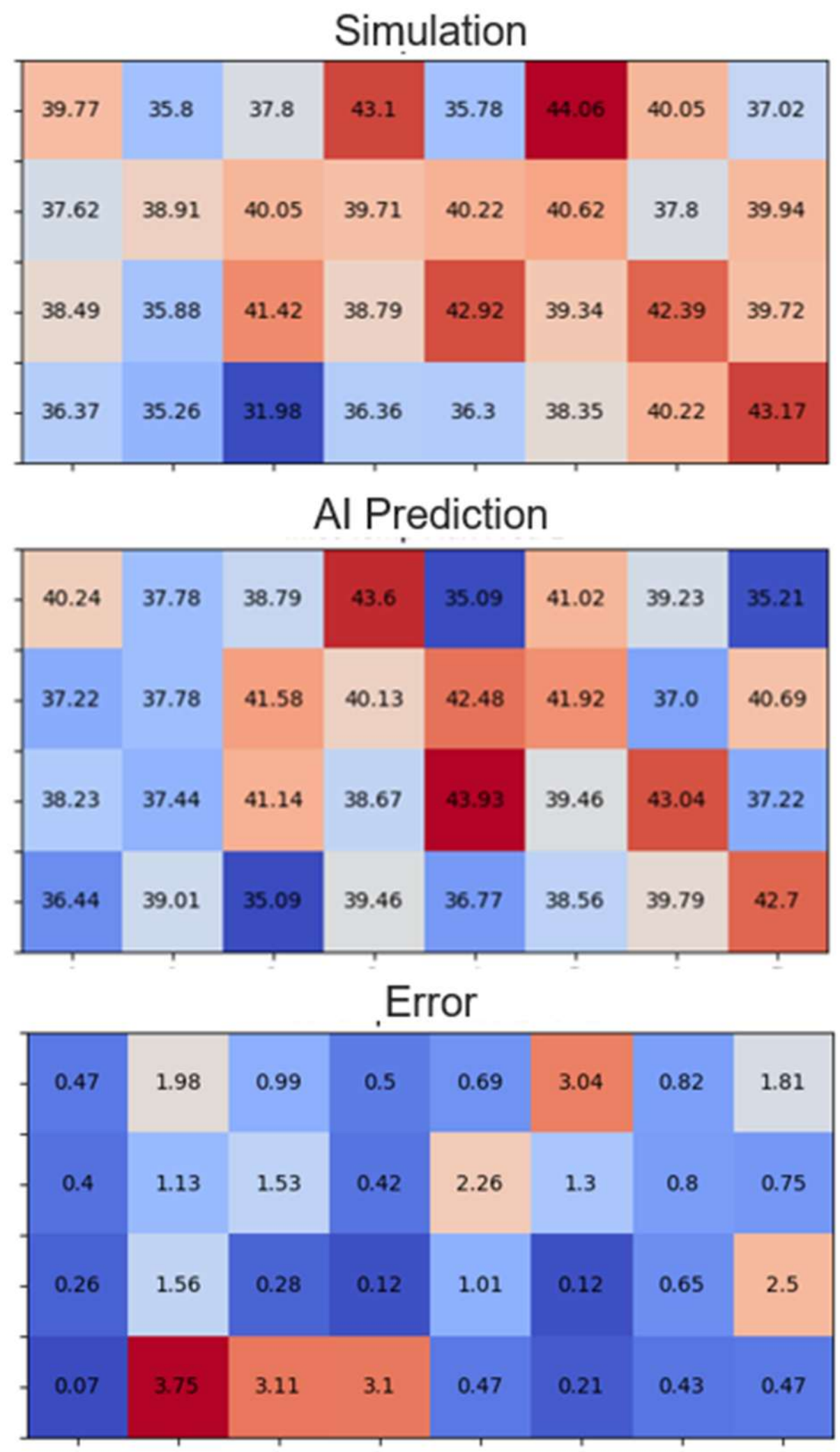

Figure 4 a) Simulation results of max cabinet inlet temperature, b) AI prediction of the raised floor design at the cabinet inlet, c) AI prediction error

### 3.3 Capacity Planning Framework

AI based prediction provides significant advantages in optimization workflows, where many design alternatives must be evaluated. The data acquisition and training effort is amortized over numerous inferences, allowing the surrogate model to be applied repeatedly at negligible computational cost, and the time savings grow substantially as the number of evaluated design alternatives increases. Consider a capacity planning scenario in which a data center has four available cabinets (M7, M8, M10 and M11), and a new workload must be assigned. Each possible placement option can be evaluated in parallel within milliseconds by the surrogate model, enabling rapid identification of the most thermally favorable configuration. To illustrate this process, both AI inference and full CFD simulation were performed for the four candidate cabinets M7, M8, M10, and M11.
As power is assigned to each candidate location, the AI model updates the predicted maximum temperature of the corresponding cabinet immediately. The comparison results are shown in Figure 5, and quantitative accuracy is summarized in Table 2. Although the absolute prediction error of the maximum temperature ranges from approximately 1 C to 3 C, the model consistently preserves the relative ordering of maximum cabinet inlet temperature across scenarios. Specifically, both AI and CFD simulations indicate the placement candidate that assign the workload to M7 produces the highest maximum temperature, followed by candidate that assign placement to M8, M10, and M11. This ability to correctly rank placement options provides practical guidance for optimization and supports efficient decision making in capacity planning workflows.

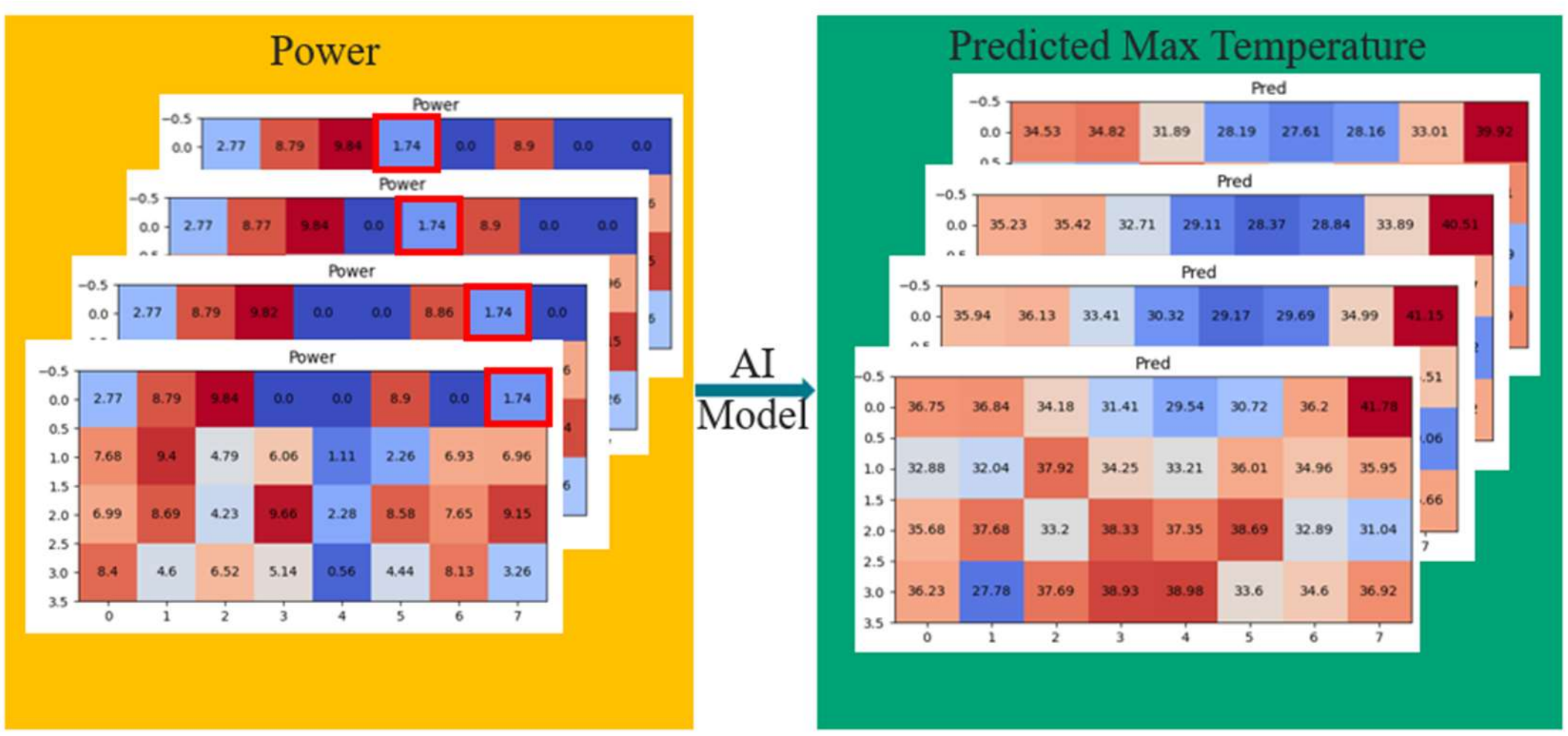

Figure 5 Maximum cabinet inlet temperature for different workload placement candidates

| Insert cabinet | Simulated max T (C) | Predicted max T (c) | Error (C) |
|---|---|---|---|
| M7 | 39.92 | 36.64 | 3.28 |
| M8 | 40.51 | 38.57 | 1.94 |
| M10 | 41.15 | 39.17 | 1.98 |
| M11 | 41.78 | 39.98 | 1.8 |

Table 2 Maximum cabinet inlet temperatures for different workload placement candidates.

## 4. AI-Driven Capacity Planning Framework

In this work, we present not only an AI driven framework for accelerating data center thermal prediction, but also a holistic, fully automated solution for data center design using AI based physics modeling. The end-to-end workflow includes data acquisition, model training, and real time inference, each step is fully automated and the entire workflow is implemented within the Cadence Reality Digital Twin platform. This integration enables users to transition seamlessly from simulation to surrogate modeling without manual intervention or external scripting.

For data acquisition, we provide a user configurable interface that allows engineers to specify the parameters to vary, their corresponding ranges, and the number of simulation cases to generate. Once configured, simulation jobs can be executed either locally or dispatched to a compute cluster through LSF. Key job-level metrics, such as the maximum temperature across the data center, are monitored in real time while simulations are running.

For training, users can likewise submit model training tasks through LSF and track progress by monitoring key metrics such as mean absolute error and maximum absolute error at each epoch. After training completes, the resulting AI model can be used directly within the Digital Twin environment for fast evaluation.

For inference, once a user loads both a data center design and a trained AI model, any modification to design parameters at the IT equipment level, rack level, ACU level or room layout level triggers an immediate inference of cabinet inlet temperatures. As illustrated in Figure 6, results are rendered instantly in the 3D environment, enabling true drag and drop exploration of different configurations. This capability provides near instantaneous thermal feedback that supports both interactive design studies and rapid capacity planning decisions.

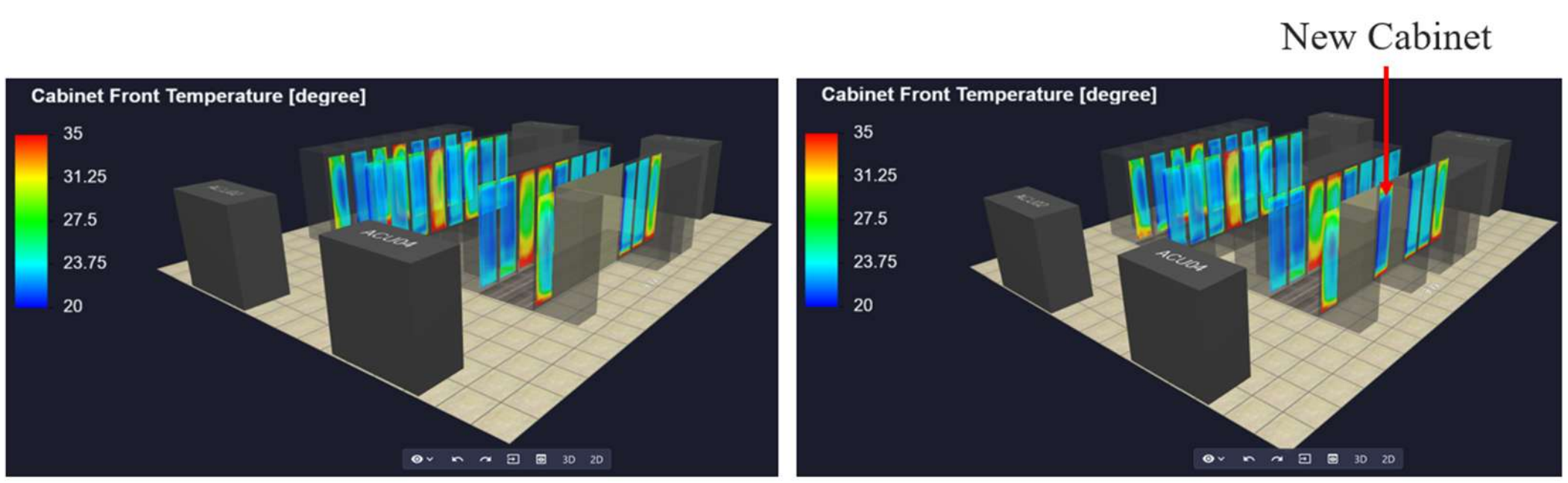

Figure 6 AI driven capacity planning framework. a) Baseline inference results, b) Inference results after assigning a new cabinet.

This fully automated platform provides significant productivity and usability benefits. By integrating data acquisition, model training, and inference into a single unified workflow, the framework eliminates the cumbersome and error-prone manual steps traditionally required for building surrogate models. Users no longer need to script simulation batches, manage data extraction, or configure training environments. Instead, they simply provide a baseline design and define the configuration parameters of interest. The platform automatically generates the simulation dataset, trains the AI model, and deploys it for real-time inference. This streamlined workflow significantly reduces both effort and expertise requirements, allowing engineers to focus on design exploration and decision making rather than managing computational pipelines.

## 5. Conclusion and Future Work

This work demonstrates a practical and scalable approach to thermal-aware data center capacity planning by integrating data generation and AI based surrogate modeling within an automated digital twin workflow. Through the development of a U-Net based surrogate model trained on a dataset of high-fidelity CFD simulations, the proposed framework achieves near real-time prediction of cabinet inlet temperatures with substantial accuracy and runtime efficiency. Across multiple data center configurations, the surrogate model delivers millisecond level inference and more than 100,000 times speedup relative to full scale CFD, while keeping reasonable accuracy and preserving the relative temperature ranking essential for workload placement. The integration of the surrogate model into the Cadence Reality Digital Twin platform further enables seamless drag and drop exploration of design variants, automated data acquisition and training, and rapid evaluation of optimization scenarios. These capabilities allow users to explore operating conditions, investigate cooling strategies and make placement decisions with unprecedented speed.

The results indicate that AI driven surrogates can support both design stage analysis and live operational decision making, reducing reliance on repetitive CFD simulations and enabling more agile management of thermally dynamic environments such as LLM hosting facilities.

Future extensions of this work include developing a foundational thermal model that generalizes across a broad range of data center architectures and can be efficiently adapted to new designs through transfer learning. Additional directions include extending the framework to predict transient thermal behavior under time-varying workloads and integrating multi objective optimization to jointly address thermal performance and energy efficiency. Through these advancements, physics AI has strong potential to become an instrumental tool for next generation data center design and operations.

## Acknowledgement

The authors want to extend our deepest gratitude to **Ziyue Liu** for his invaluable contribution.